\documentclass{article}

\usepackage{times}
\usepackage{graphicx} 
\usepackage{subfigure}
\usepackage{natbib}
\usepackage{multicol} 
\usepackage{blindtext} 
\usepackage{natbib} 
\usepackage{algorithm}
\usepackage{algorithmic}
\usepackage{amssymb}
\usepackage{amsfonts}
\usepackage{booktabs}

\usepackage{hyperref}

\usepackage[accepted]{icml2019}
\usepackage{amsmath}
\newcommand{\norm}[1]{\left\lVert#1\right\rVert}
\newcommand{\I}[0]{\mathcal{I}}  
\newcommand{\C}[0]{\mathcal{C}}
\newcommand{\Z}[0]{\mathcal{Z}}

\icmltitlerunning{Video Content Swapping Using GAN}

\begin{document}

\twocolumn[
\icmltitle{Video Content Swapping Using GAN}



\icmlsetsymbol{equal}{*}

\begin{icmlauthorlist}
\icmlauthor{Tingfung Lau (tingfunl)}{cmu}
\icmlauthor{Sailun Xu (sailunx)}{cmu}
\icmlauthor{Xinze Wang (xinzew)}{cmu}
\end{icmlauthorlist}

\icmlaffiliation{cmu}{Carnegie Mellon University, Pittsburgh, PA 15213, USA}
\icmlcorrespondingauthor{}{}

\icmlkeywords{}

\vskip 0.3in
]





\section{Introduction}
Video generation is an interesting problem in computer vision. It is quite popular for data augmentation, movie special effects, AR/VR and so on. With the advances of deep learning, many deep generative models have been proposed to solve this task. These deep generative models provide a way to utilize all the unlabeled images and videos online, since it can learn deep feature representations with unsupervised manner. And these models can also generate different kinds of images, which have great value for visual application. However, video generation still has long way to go. Directly generating a video from scratch could be very challenging since we need to model not only the appearances of objects in the video but also their temporal motion. 

There are three main challenges for video generation \cite{mocogan}. Firstly, the system needs to build appearance models and physical motion models to get both the time and space information. Any model that doesn't work well will lead to the generated video containing objects with physically impossible motion. Secondly, even when objects perform some basic motion, there are so many variations considering time dimension, such as speech. Thirdly, motion artifacts are particularly perceptible for human beings.

In this project, we focus on a particular sub-task of video generation: video content swapping. Since visual signals in a video can be divided into content and motion, which represent objects and their dynamics respectively. The goal of our project is to swap the appearance object in an video but keep the motion being the same. For example, if we have a video clip of a professional dancer performing ballet, we would like to generate video of ourselves performing a fantastic ballet show, with the same pose as the professional dancer in the original video. Users without professional knowledge can also create video with the content as they want. All they need is an original video, and a photo including the content that they want to replace. We want to develop such a system based on some conditional generative models.

This project requires to solve three major problems. First of all, we need to train a constrained generative model to make sure the output is corresponding to the input objects. It is not enough to randomly generate a normal video. Secondly, the system should recognize between foreground and background images of the video. And it can automatically adjust the screen according to the interactions between objects and context, such as occlusion. Thirdly, with the increase of the number of people in the image, the system can still maintain real-time performance. \cite{cao2018openpose}

A wide range of generative methods having been proposed for image generation, including Variational Autoencoders (VAE) \cite{kingma2013auto}, Generative Adversarial Networks (GAN) \cite{NIPS2014_5423}, and Conditional Generative Adversarial Networks (CGAN) \cite{mirza2014conditional}. 
For video generation, there are several approaches using GAN, such as Temporal GANs conditioning on Captions (TGANs-C) \cite{pan2017create}, Motion and Content decomposed Generative Adversarial Network (MoCoGAN) \cite{mocogan}, and Semantic Consistent Generative Adversarial Network (SCGAN) \cite{yang2018pose}. \cite{denton2017unsupervised} propose an alternative way using disentangled auto encoder.

As we focus on video content swapping, our video generation should be conditioned on content input and pose in a reference video. 
We first extract the pose information from a video using a pre-trained human pose detection \cite{cao2018openpose} trained on large scale data set.
Then we will use a generative model to synthesize a video using the extracted pose and content image.
This could be done using a disentangled auto encoder that learns to reconstruct an image based on content code and pose code or using a cGAN that conditioned on pose and content image.
We will do experiments with these two methods on 3 data sets to find out the best method for solving this task. 
We plan to make up a tool for user to create their own synthesized video based on our method.

\begin{figure*}[htbp]
    \centering
    \includegraphics[width=0.9\linewidth]{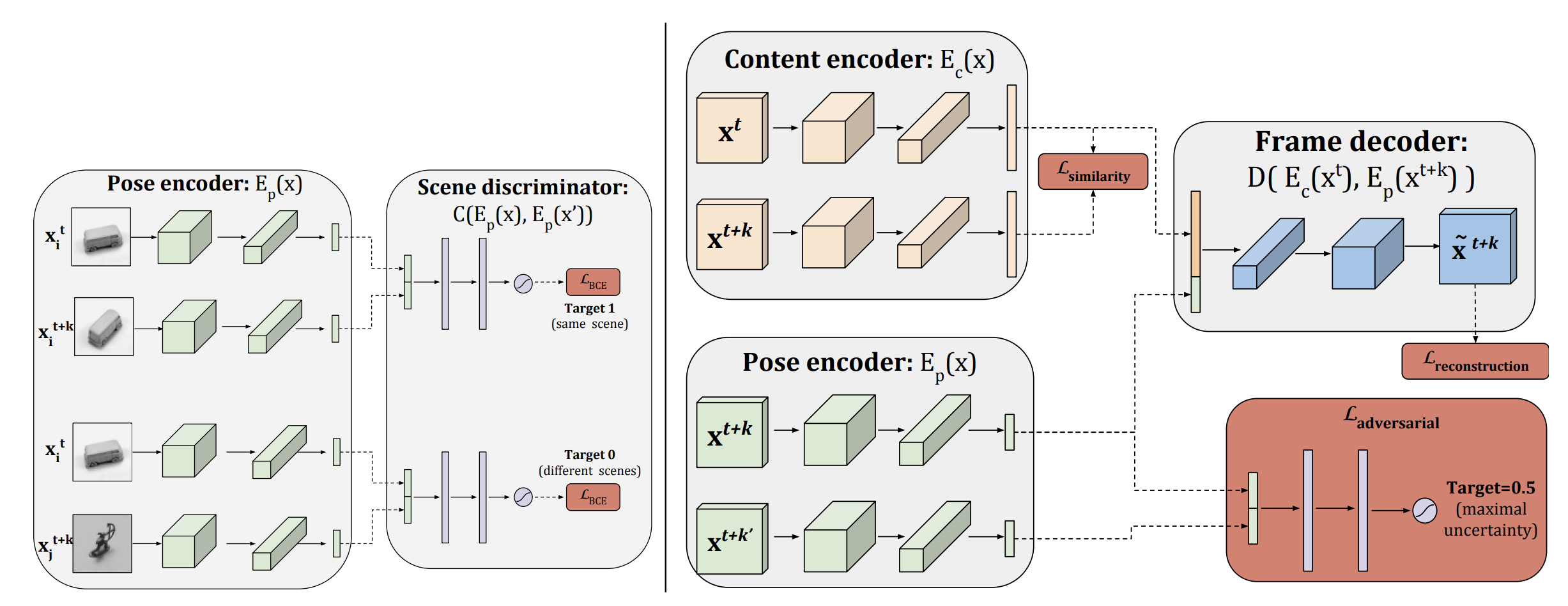}
    \caption{Architecture for Video Generation}
    \label{fig:video_gener}
\end{figure*}

\section{Related Works}

\subsection{Video Content Swapping.} The swapping problem in videos, specifically human faces, has been studied via various approaches. Traditional methods such as \citep{f2f} and \citep{daleface} deliver impressive results, and the former one even offers real-time video reenactment. However, these methods rely heavily on domain-specific knowledge and fine-tuned features specifically designed for face, and thus could not be generalized for non-face tasks. 

A recent unpublished work based on deep network called DeepFake takes on this problem by training a shared encoder, and decoding using the target-specific decoder. This approach also gives impressive results and does not require domain-specific knowledge. However, it requires large number of training samples of the target face. Even worse, we need to train a different encoder for every different target face that we want to map to.

\subsection{Generative Model}
A huge amount of deep generative models have been proposed for image generation and video generation. There are two basic structures for these models, which is VAE and GAN. \citet{kingma2013auto} proposed a stochastic variational inference and learning algorithm, which can be applied to large datasets and intractable cases efficiently, \cite{rezende2014stochastic} introduced a recognition model to represent
an approximate posterior distribution and uses this for optimisation of a variational
lower bound. \cite{tulyakov2017hybrid} presented a principled framework, which is Hybrid VAE (H-VAE), to capitalize on unlabeled data.
\citet{NIPS2014_5423} proposed GAN to solve image generation by learning a generator to fool a discriminator for classifying whether image is real/fake. 
\citet{mirza2014conditional} further extend the idea and proposes conditional generative adversarial nets (CGAN). 
The generator models the conditional distribution of data
and the discriminator takes the variables conditioned on as an
additional input. 
We also use the same idea of CGAN in this project.

\subsection{Video Generations} \citet{denton2017unsupervised} proposes a different idea by taking into account the temporal coherence of a video, albeit not for the face-swapping task but for video prediction task. It disentangles the video representation into two components--\textit{content}, which remains unchanged throughout a single clip or a duration of time, and \textit{pose}, which captures the dynamic aspects of the clip and thus varies over time. Concretely, for the pose constraint, they introduce the adversarial loss for a discriminator network $C$ and pose encoder $E_p$, and the $i$-th clip at $t$'s frame $x_i^t$:

\begin{gather*}
\mathcal{L}_{adversarial}(C) = -\log(C(E_p(x_i^t), E_p(x_i^{t+k}))) \\
-\log(C(E_p(x_i^t), E_p(x_j^{t+k}))) \\
\mathcal{L}_{adversarial}(E_p) = -\log(C(E_p(x_i^t), E_p(x_i^{t+k}))) \\
-\log(C(E_p(x_i^t), E_p(x_i^{t+k})))
\end{gather*}

Effectively, the loss encourages the pose encoder to produce pose embedding that is indistinguishable for a discriminator. The whole architecture is shown in Figure \ref{fig:video_gener}. The discriminator in the left part is trained with binary cross entropy (BCE) loss to predict whether a pair of pose vectors comes from the same (top portion) or different (lower portion) scenes. When the discriminator is trained, the pose encoder $E_{p}$ is fixed. The right part is the whole model. And when the pose encoder $E_p$ is updated, the scene discriminator
is held fixed.

\citet{yang2018pose} divides the human video generation process into two steps, the first step is to synthesize the human pose sequence from an initial pose at the first frame, the second step is to generate the video frames from an human pose sequences and initial frame of the video.
They train their pose generator and video generator using the GAN loss to classify whether the generated pose sequences and video are real or fake.

\citet{mocogan} also proposes the idea of replacing the video entity by swapping the content embedding. However, their content embedding is sampled once and assumed fixed for the entirety of the clip, which is overly constrained as there might be object occulation and rotation in the video, which might change the content embedding. Moreover, they could not specify a target content as they do not have the encoder part, as they are only relying on the unconditional distribution of the latent content embedding.

\subsection{Pose Extraction}\label{subsec:pose_extraction}
Pose extraction is an important problem for a wide range of scenarios, and people have proposed and optimized different kinds of models for different situations. In the multiple person pose estimation, there are two types of methods, which are part based and two steps. In the part-based framework, the model first detect all body parts, and then label and assemble these parts to detect the pose. But since only small local regions are considered, the body-part detectors are vulnerable. 

\begin{figure}[H]
    \centering
    \includegraphics[width=0.9\linewidth]{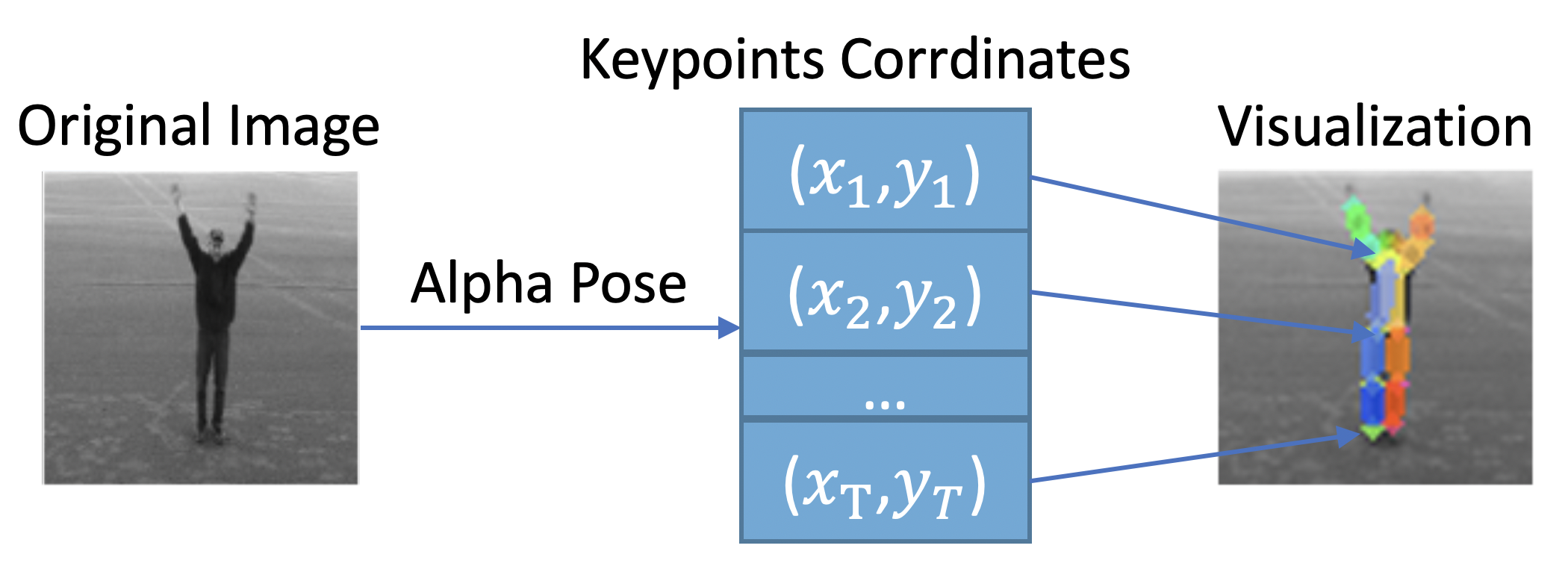}
    \caption{Architecture for Pose Extractor}
    \label{fig:alphapose}
\end{figure}

The two-step framework extracts pose based on object detection and single person pose estimation, and it has better performance. In this project, we utilized AlphaPose \cite{fang2017rmpe}, which follows two-step framework. The whole process can be divided into three steps, firstly, the human detector bounds humans in the images with boxes. Secondly, these boxes are fed into the 'Symmetric STN' and 'SPPE' module, and the pose proposal are generated automatically. At last, the generated pose proposals are refined by parametric Pose NMS to obtain the estimated human poses. Especially, AlphaPose introduces 'Parallel SPPE' to avoid local minimums and leverage the power of SSTN in the training process. And designing a pose-guided proposals generator to augment the existing training samples. This method have better performance for multi-person human pose estimation in terms of accuracy and efficiency. Therefore, we extract human poses in videos with AlphaPose, and the process is shown in Fig.\ref{fig:alphapose}.

\section{Methods}
\subsection{Disentangled Representation}\label{sec:disentangled_repr}
Inspired by \citet{denton2017unsupervised}, we want to solve this problem by learning a disentangled representation of video. 
The idea is that we use a auto-encoder like model to encode a video $\{\mathcal{I}\}_{t=1}^{T}$ to a content code $\{\mathcal{C}\}_{t=1}^{T}$ and a pose code $\{\mathcal{Z}\}_{t=1}^{T}$ and generate the video from the hidden code. We denote the $t$-th frame of the $i$-th video by $\mathcal{I}_i^t$, and similarly for all other sequential collection of variables.
The model have a pose encoder $F_{pose}$ and a content encoder $F_{content}$ and a generator $G$, the hidden code and reconstruction is computed as 
\begin{align*}
    \mathcal{C}_i^t & = F_{\mathit{content}}(\mathcal{I}_i^t) \\
    \mathcal{Z}_i^t &= F_{\mathit{pose}}(\mathcal{I}_i^t) \\
    \hat{\mathcal{I}}_i^t & =  G(\mathcal{C}_i^t, \mathcal{Z}_i^t).
\end{align*}
for each time frame $\mathcal{I}_i^t$ in a video sequence. 

We want to ensure the learned content code and the pose code captures exactly the appearance and motion features respectively.
For the pose code, since we focus on video with a human as its subject, so a natural idea is to utilize the off-the-shelf deep human pose estimators pretrained on large scale image data sets \citep{cao2018openpose}, as already mentioned in section \ref{subsec:pose_extraction}. Therefore the $F_{\mathit{pose}}$ would be a fixed pre-trained model throughout the training process. We could first extract $\mathcal{Z}_i^t$ in one pass in the preprocessing step and use it for later training.

For the content code, intuitively the content code should not vary much throughout the  consecutive frames. So we use the following consistency loss to ensure the content encoding of frame $t$ and $t+k$ for $k$ random sampled from some time-window $[-w,w]$ is consistent\citep{denton2017unsupervised}.
\begin{equation}
    \mathcal L _{\mathit{consist}} = \Vert F_{\mathit{content}}(\mathcal{I}_i^{t+k}) - F_{\mathit{content}}(\mathcal{I}_i^{t}) \Vert_2^2
\end{equation} 

As a standard measurement of the quality of decoder and encode, we compare the self-reconstructed frame $\hat{\I_i^t}=G(\C_i^t, \Z_i^t)$ against the original frame $\I_i^t$: 
\[ L_{rec} = \norm{\I_i^t - \hat\I_i^t}_2^2\]

\begin{figure}[htbp]
    \centering
    \includegraphics[width=\linewidth]{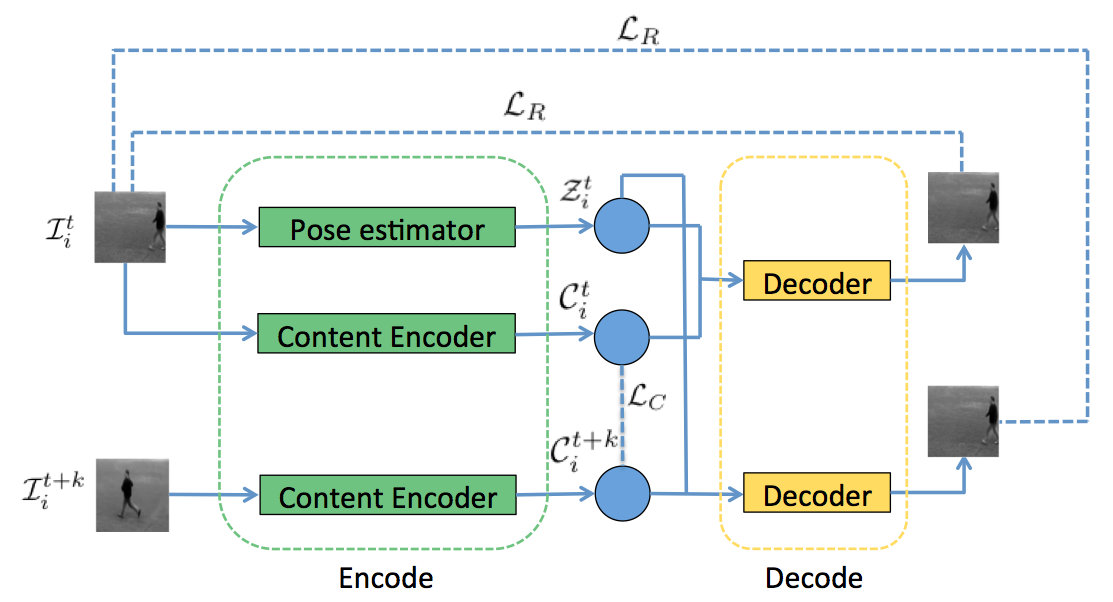}
    \caption{The whole pipeline of disentangled video representation learning.}
    \label{fig:pipeline}
\end{figure}

Apart from the standard self-reconstruction error, according to our assumption, if we use the content code at time $t+k$: $\mathcal C_{i}^{t+k}$, and the pose code at time $t$: $\mathcal Z_i^{t}$ to reconstruct the image, the image should be close to $\mathcal I_i^{t}$, therefore we impose the following temporal-shifted reconstruction loss on our encoder-decoder network as in \citet{denton2017unsupervised}.
\begin{equation}
    \mathcal L_{\mathit{rec}} = \Vert \mathcal{I}_i^{t} - G(F_{content}(\mathcal{I}_i^{t+k}), \mathcal{Z}_i^{t})) \Vert_2^2 
\end{equation}
The whole model is shown in Figure \ref{fig:pipeline}. It could be trained on human video data set to learn the disentangled representation.

During inference time, we first compute the pose codes $\{\mathcal{Z}_i\}_{t=1}^T$ for a video using the pretrained human pose encoder $F_{\mathit{pose}}$ and then compute the content code $\Tilde{\mathcal{C}}$ for some an reference image for content swapping.
Then we use the generator to produce images sequence $\Tilde{\mathcal I}^t = G(\Tilde{\mathcal{C}}, \mathcal{Z}_i^t)$ as the edited video.
Since the encode disentangle the content and pose, so the resulting system will be able to achieve the goal of video content swap.

The content encoder in the original work is a shallow 5 layer convolutional network to produce features at different resolutions.
The decoder uses 5 layer transposed convolution and combines hidden code at different resolutions to generate a image.
The structure of the network is describe in Section \ref{sec:enc-dec}. 

\subsection{Conditional GAN}
GAN is a popular generation model, however, simply using GAN cannot solve the video content swapping problem because we require the output being conditioned on both the content and pose image. \citet{mirza2014conditional}'s work has inspired us to come up with our conditional GAN framework. We will train a generator $G$ that generates a frame conditioned on a content code $\I$ and a pose code $\Z$. The generator is the same encoder-decoder structure as we use in previous approach, but we optimize the network using the GAN loss. 
Since we condition the generation upon two hidden codes, we train two separate discriminators, the pose discriminator $D_{\mathit{pose}}$ and content discriminator $D_{\mathit{content}}$.
The pose discriminator takes a frame $\I_i^t$ and a corresponding pose $\Z_i^t$ to classify whether this is a true pair from the data or a frame generated by generator conditioned on the pose code $Z$. This give us the loss of pose discriminator
\begin{align*}
    \mathcal{L}_{\mathit{pose}} &= \mathbb{E}[-\log D_{\mathit{pose}}(\mathcal{I}_i^t | \mathcal{Z}_i^t)] \\
    & \quad + \mathbb{E}[-\log (1- D_{\mathit{pose}}(G(C_i^t, \mathcal{Z}_i^{t'}) | \mathcal{Z}_i^{t'}))]
\end{align*}
The pose coordinates are converted to a heat map with the same size as the input image, before putting into the discriminator.
Specifically speaking, we use 2D Gaussian distributions $\mathcal{N}((x_i, y_i)^{T}, \sigma^2 I)$ to encode human key points $(x_i, y_i)$ for $i=1,\cdots, 17$. The Gaussian distribution is then converted to a $17 \times 128\times 128$ feature map by calculating the density function at each grid point. The resuling feature map is concatenated with the $3 \times 128 \times 128$ as the input.

We also train a discriminator $D_{\mathit{content}}$ to ensure the output looks like real image and has the same appearance as the content image.
$D_{\mathit{content}}$ will take a pair of image, and decide whether it is two real samples from the same video, or a generated frame and the content image the generator conditioned on. This give us the loss of content discriminator
\begin{align*}
    \mathcal{L}_{\mathit{content}} &= \mathbb{E}[- \log D_{\mathit{content}}(\mathcal{I}_i^t, \mathcal{I}_i^{t'})] \\
    & \quad  + \mathbb{E}[-\log (1 - D_{\mathit{content}}(G(\mathcal{I}_i^t,\mathcal{Z}_i^{t'}),\mathcal{I}_i^{t}))
\end{align*}
The generator optimize the GAN loss 
\begin{align*}
    \mathcal{L}_{\mathit{GAN}} &=  \mathbb{E}[-\log (D_{\mathit{content}}(G(\mathcal{I}_i^t,\mathcal{Z}_i^{t'}),\mathcal{I}_i^{t}))] \\
    & \quad + \mathbb{E}[-\log (D_{\mathit{pose}}(G(I_i^t, \mathcal{Z}_i^{t'}), \mathcal{Z}_i^{t'}))]
\end{align*}

\begin{figure*}[htbp]
    \centering
    \includegraphics[width=\linewidth]{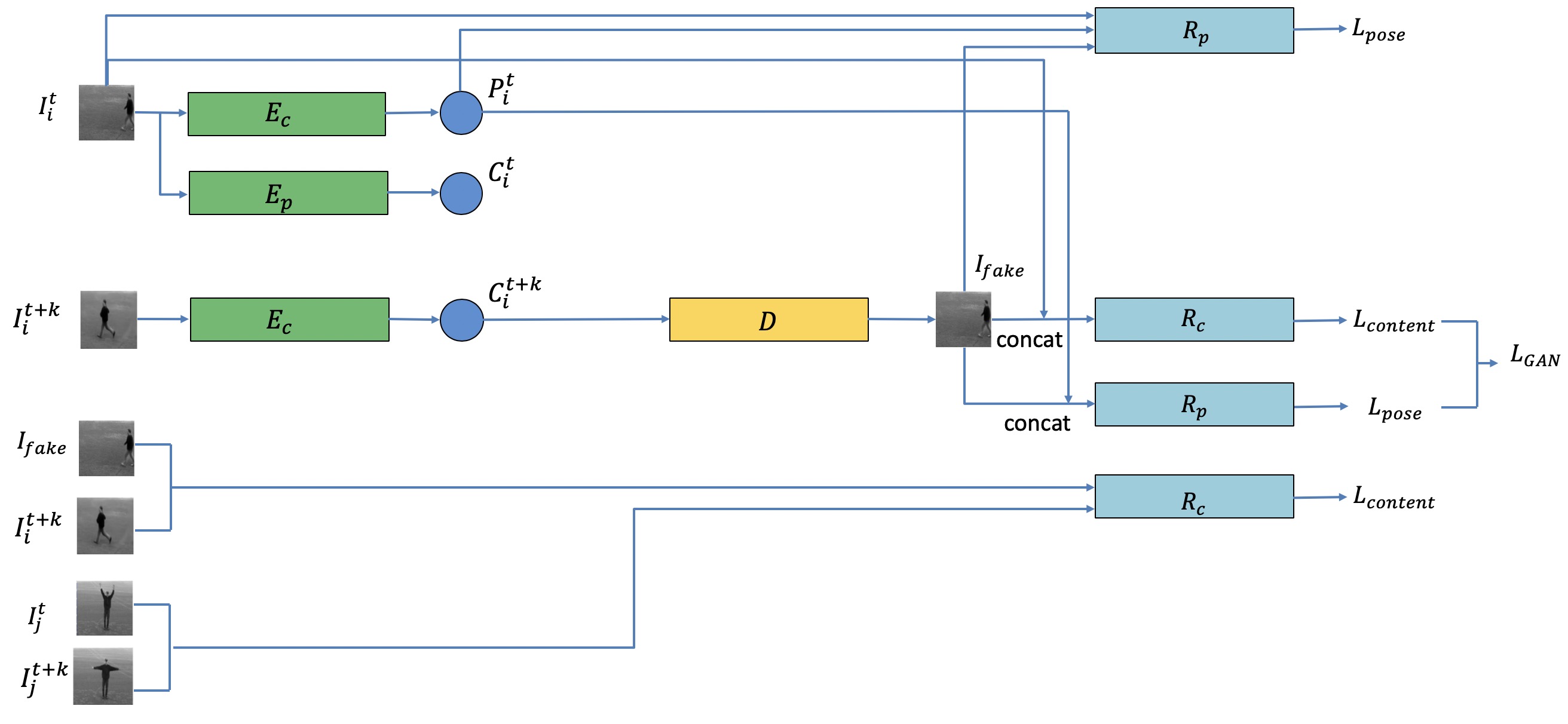}
    \caption{The whole pipeline of our proposed Conditional GAN model.}
    \label{fig:cgan_pipeline}
\end{figure*}

The whole model could be concluded by Figure \ref{fig:cgan_pipeline}

\paragraph{Triplet Content Discriminator}\label{para:triplet} The content discriminator in the previous conditional GAN method decides whether two frames are from the same video. 
Our first approach simply concatenates two images and pass it through a convolutional network.
To better compare the similarity in content between two frames, we proposes another structure, that passes the original frame through the convolutional network to get a content embedding. Moreover, we make use of the triplet loss that encourages the similarity between an anchor and its positive pair and dissimilarity between the anchor and its negative pair. \\
To illustrate, in our case, the anchor and the positive content code would be two content codes from two different frames of the same video $c_\textit{anchor} \triangleq \C_i^t, c_\textit{positive} \triangleq \C_i^{t'}$, and the negative content code would be the content code from a reconstructed frame $c_\textit{negative} \triangleq F_\textit{content}(G(C_i^t, Z_i^{t'}))$ then the triplet loss is defined as 
\begin{equation}
    \max(0, \Vert c_{\textit{anchor}} - c_{\textit{positive}} \Vert^2
    + m - \Vert c_{\textit{anchor}} - c_{\textit{negative}} \Vert^2).
\end{equation}
where $m$ is the margin hyperparameter, by which we want different classes to be separated 

\subsection{Encoder and Decoder}
\label{sec:enc-dec}
\begin{figure}[htbp]
    \centering
    \includegraphics[width=\linewidth]{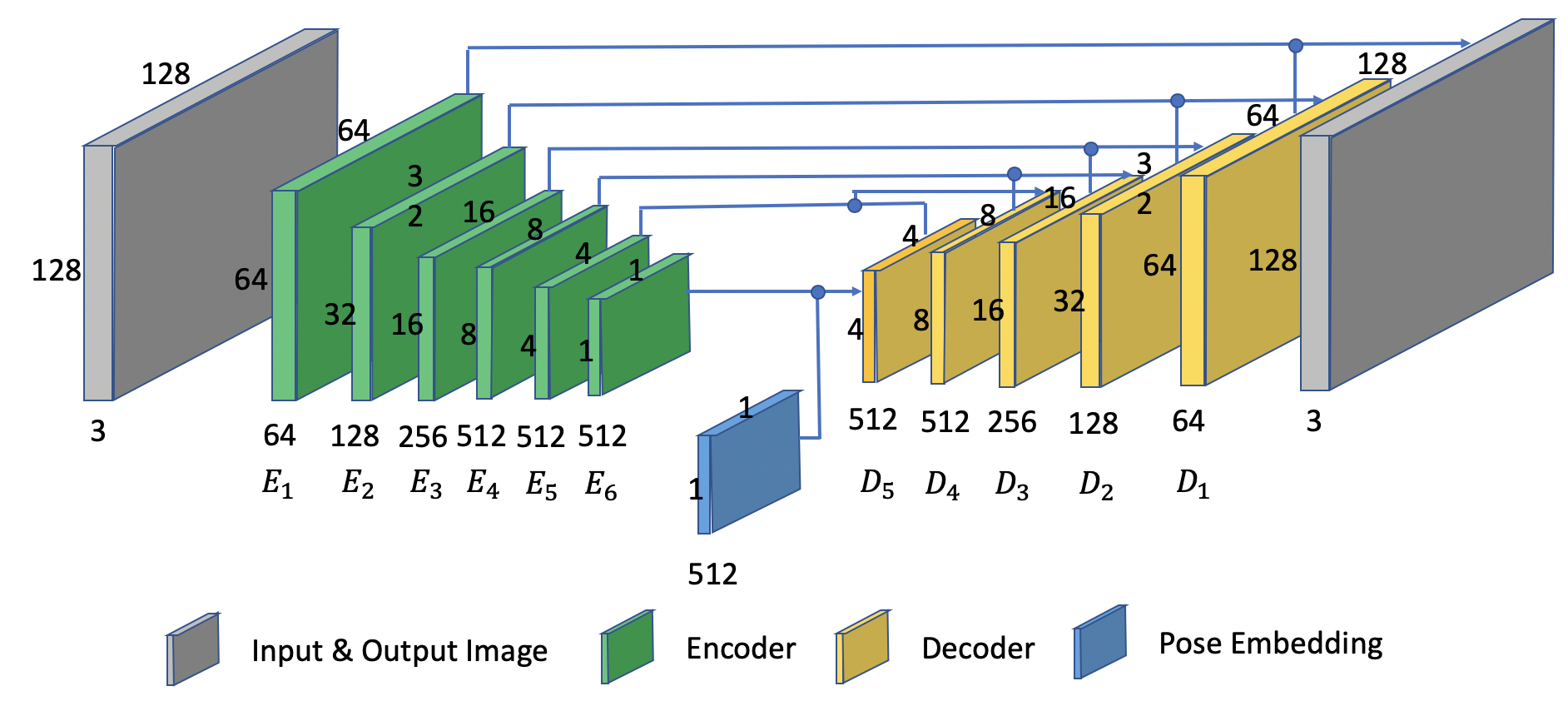}
    \caption{The Architecture for encoder and decoder}
    \label{fig:en_de}
\end{figure}

We build the encoder and decoder as convolutional models, both of which contain six convolutional layers. The architecture is shown in Fig.\ref{fig:en_de}. 
Using more complex models like adding additional convolutional layers or using residual blocks as in \citet{zhu2017unpaired} may further improve the capacity of the generator. 
However, this project is focused on the probabilistic formulation of the video frame generation task, and this structure is effective enough for our task. Thus we stick on this structure.

\section{Experiment}
\subsection{Data}
The most important point for the dataset is that the videos and images should contain people in motion. Since in encoding process, pose estimator would extract the pose of people based on pretrained human pose frame. It will be difficult to extract the pose of other moving objects. Therefore, we searched for several datasets containing moving people to evaluate our model from different aspects. We have also searched for some other datasets, however, many videos contains more than one subject and the number of subjects varies in a single video from one frame to another. We therefore adhere only to the two datasets  below for all our experiments.

\begin{itemize}
    \item \textbf{KTH Action.} \footnote{http://www.nada.kth.se/cvap/actions/}
    \begin{figure}[htbp]
        \centering
        \includegraphics[scale=0.4]{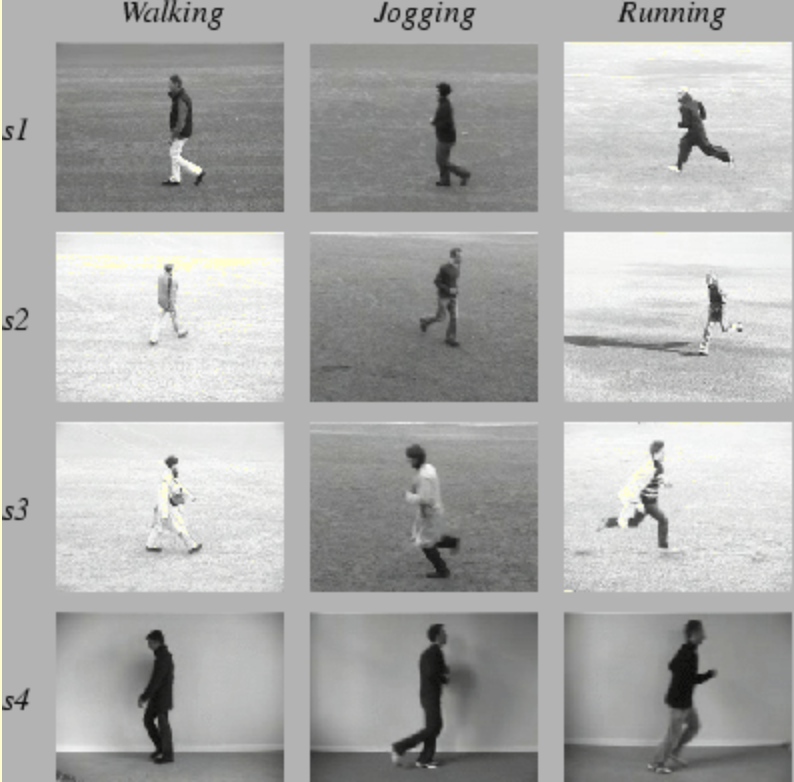}
        \caption{KTH Action Samples}
        \label{fig:kth_sample}
    \end{figure}\\
    KTH Action is one of the most popular datasets for recognition of human actions. This video database contains six types of human actions, which are walking, jogging, running, boxing, hand waving and hand clapping. These actions are performed several times by 25 subjects in four different scenarios: outdoors $s_1$, outdoors with scale variation $s_2$, outdoors with different clothes $s_3$ and indoors $s_4$. Currently the database contains 2391 sequences. All sequences were taken over homogeneous backgrounds with a static camera with 25fps frame rate. The sequences were down-sampled to the spatial resolution of $160 \times 120$ pixels and have a length of four seconds in average.

    \item \textbf{UCF101.} \footnote{https://www.crcv.ucf.edu/data/UCF101.php} 
    \begin{figure}[htbp]
        \centering
        \includegraphics[scale=0.4]{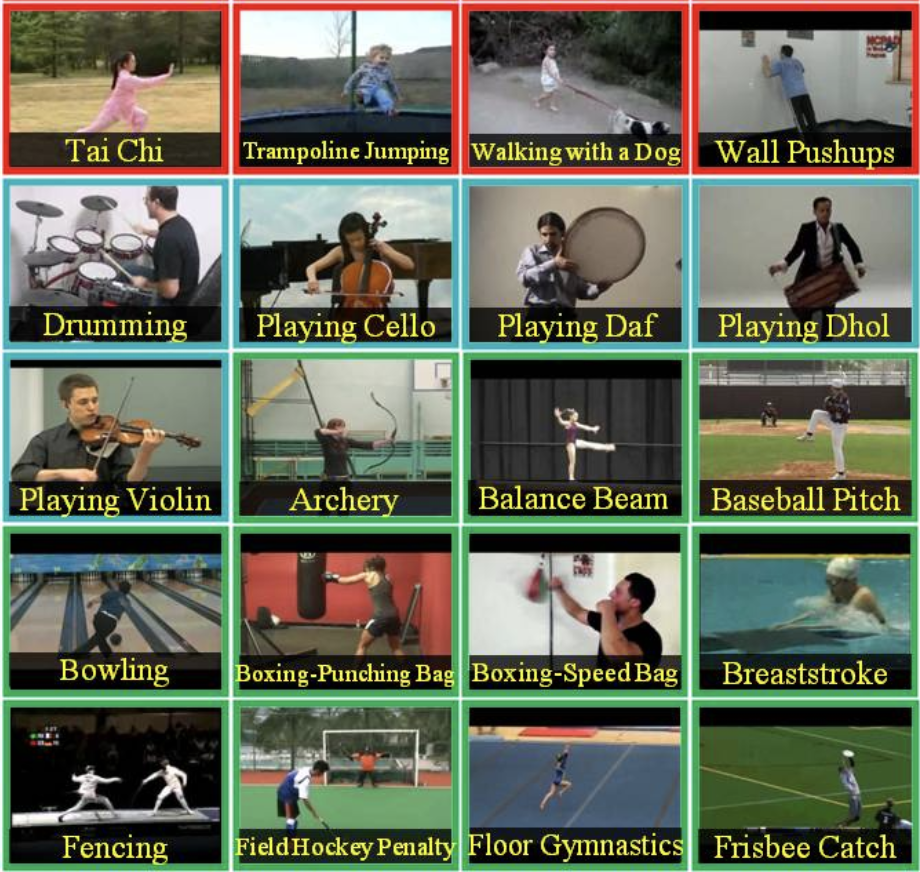}
        \caption{UCF-101 Samples}
        \label{fig:ucf101_samples}
    \end{figure}   
    This dataset is commonly used for video action recognition. It includes 13,220 videos of 101 different action categories, such as basketball shooting, biking/cycling, diving, golf swinging and so on. This data set is very challenging due to large variations in camera motion, object appearance and pose, object scale, viewpoint, cluttered background, illumination conditions, etc. For each category, the videos are grouped into 25 groups with more than 4 action clips in it. The video clips in the same group share some common features, such as the same actor, similar background, similar viewpoint, and so on.
\end{itemize}


\begin{table*}[htbp]
    \centering
    \begin{tabular}{ccccc}
    \toprule
       Method  &  Disentangled Baseline & Disentangled with Pretrained Pose & CGAN & CGAN with Triplet Loss  \\ \midrule
       MSE Loss & 0.0036 & 0.0040 & 0.0048 & 0.0061 \\
       \bottomrule
    \end{tabular}
    \caption{Reconstruction mean squared error of the 4 methods.}
    \label{tab:mse}
\end{table*}

\begin{figure*}[htbp]
    \centering 
    \includegraphics[width=\textwidth]{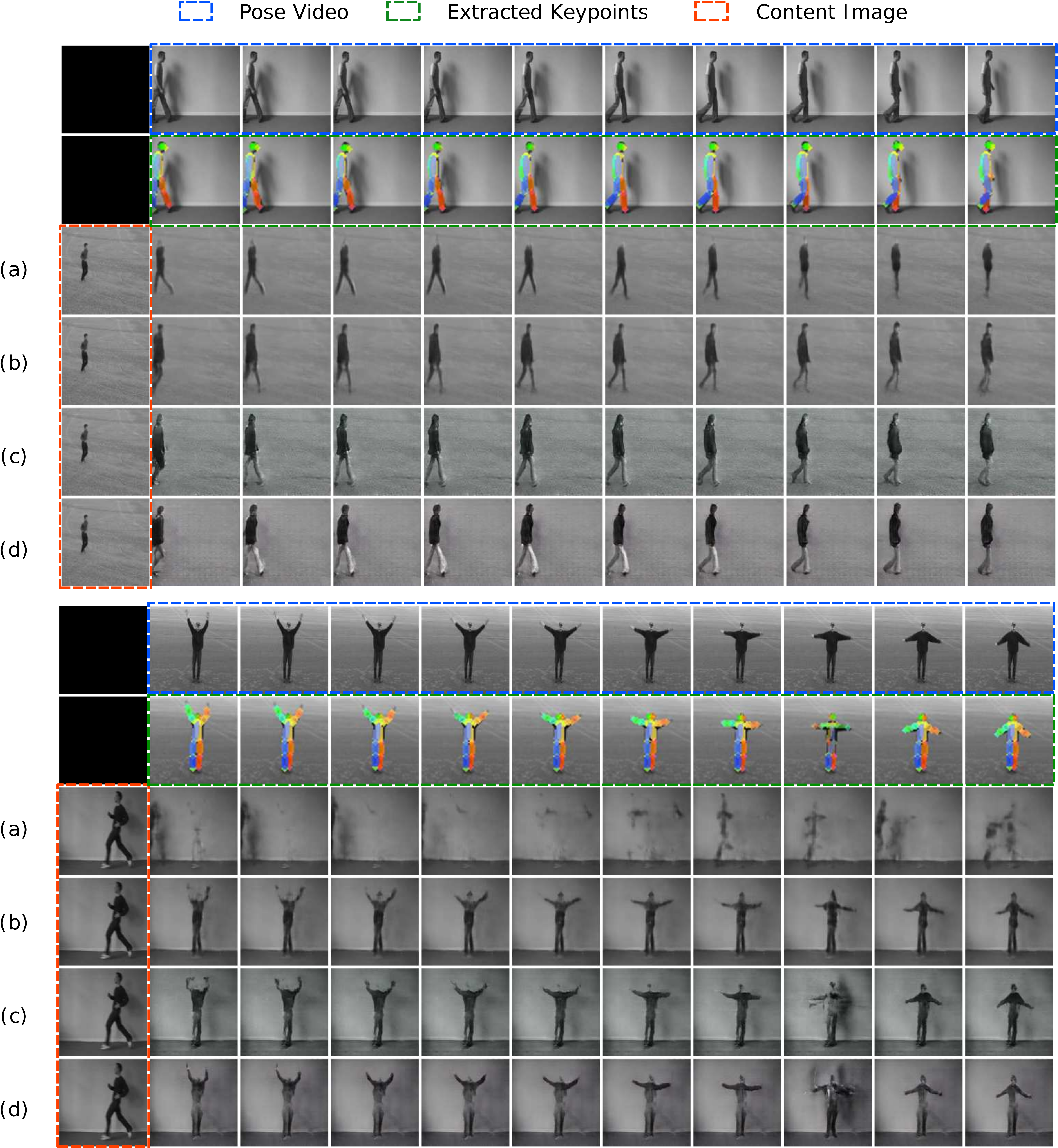}
    \caption{Visualization of video content swap results with 4 methods. (a) Disentangled Baseline (b) Disentangled with Pretrained Pose (c) CGAN (d) CGAN with Triplet Loss. }
    \label{fig:content-swap}
\end{figure*}

For some datasets, the scaled videos are really small, so quantitative evaluation may not faithfully reflect the quality of the generation. Therefore, we will provide more visual results in the sequel sections.

\subsection{Pose Extraction}
We preprocess each video with the AlphaPose \cite{fang2017rmpe}. The return format is a list containing 2-D coordinates of 17 key points: $(x_i,y_i)_{i=1}^M$, where $i$ denotes the $i$-t h key-point, e.g: head, hands, and etc.
The coordinate is normalized to range $-1$ to $1$ with respect to the image width and height.
In some image the human is fully or partially outside the image, the pose detector will not detect the pose in this case. 
To tackle this problem, we also add one binary variable to indicate whether the human is visible. Therefore we have a 35 dimension vector in total per person per frame. 
For the sake of simplicity, we will be dealing with video with single subject (human).


\subsection{Experiment Details}
For a fair comparison of four different methods, including the baseline and ones we proposed above:
\begin{itemize}
    \item[(a)] Disentagled Baseline
    \item[(b)] Disentangled with Pretrained Pose
    \item[(c)] CGAN
    \item[(d)] CGAN with Triplet Loss
\end{itemize}
We implemented these methods based on the original implementation \footnote{https://github.com/edenton/drnet} of \citet{denton2017unsupervised}. We further develop the pose extraction part and the CGAN method, our code is released\footnote{https://github.com/ldf921/drnet-py}.

Due to the limitation of computation resources, we stick to the KTH dataset for all our experiments. We train each proposed method for 100 epochs, using the same optimizer configuration: Adam with default learning rate at 0.001 for all different models.

\subsection{Experiment Results}
We evaluate these different methods by the MSE reconstruction loss mentioned in \ref{sec:disentangled_repr}, as listed in  \ref{tab:mse}. For the quality of the disentangled learning, we have also visualized the result of swapping the content code of a video with the content code extracted from a different video,  in hope of seeing the very same person performing different actions. In figure \ref{fig:content-swap}, we have shown two content swapping examples. The first example has different subject magnitudes for the \textit{content image} and \textit{pose video} but poses are on the easier side, whereas the second example has roughly the same size of both subjects  yet the pose are much harder. Each example consists of four different methods. The blue row of each example shows the \textit{pose video} from which we extracted the pose code, and the green row of each example shows the key-points that we have extracted using AlphaPose and will use in method (b),(c),(d). The red column of each example shows the (single) \textit{content image} from  which we extract the content code and swap into the \textit{pose video}.

\paragraph{Disentangled Baseline}
In the first example, we can see that the network has marginally learned disentangled representation, and the swap-reconstruction look close to the expected goal.
However, the quality of generated image is low as the image is blurry, and details of body parts are obviously lacking: we could barely see the head of the person from the \textit{content image}.
In the second example, the result is much worse, we could see that the subjects from both videos get mixed up in the final result and are both blurry, which indicates that the content information is not flowing through the content code solely. Therefore the pose code will be polluted by the content information of the \textit{pose video} and the content codes of both \textit{pose video} and \textit{content image} are salient in the swap-reconstruction result. Additionally, although the MSE Loss is the lowest for this baseline method, the swap-reconstruction result is actually inferior to methods we have proposed, showing that $\ell_2$ is not a accurate indicator of the quality of the result.

\paragraph{Disentangled with Pretrained Pose}
Apparently, the fidelity of the swap-reconstructed video has significantly improved from the Baseline Disentangled Model. In the first example, we can see that the details of the subject of the \textit{content image} are well preserved in the swap-reconstruction. Moreover, as a sign of better learned disentangled representation, we no longer see the content clutter as before in the second example. Interestingly, as a result of using a pretrained pose, we see that when we have drastically different subject magnitudes in \textit{content image} and \textit{pose video} (as in the first example), the subject of the \textit{content image} will be re-sized to the magnitude of the subject of the \textit{pose video}.

\paragraph{CGAN}
CGAN as well as CGAN with triplet-loss below, has much sharper swap-reconstruction than both the Disentangled Baseline and the Disentangled with Pretrained Pose. However, in the second example, this method again suffers the problem of cluttering the content information from both \textit{content image} and \textit{pose video}, which is salient in the third to last frame.

\paragraph{CGAN with triplet-loss}
Although we have added the triplet-loss to encourage similarity between content codes of the same video and to simultaneously encourage dissimilarity between content codes of different videos with a separation of \textit{margin} as mentioned in \ref{para:triplet}, the swap-reconstruction result is no better than the CGAN--actually the reconstruction is blurrier than the ordinary CGAN in the first video. In the second experiment, the subject cluttering problem also deteriorates as we can see the subjects of both videos overlapping each other (notice a shadow of the running pose from the \textit{content image}) in every single frame. We think that further fine-tuning the \textit{margin} hyperparameter could fix the problem, or maybe a $\ell_2$ norm in the representational space is not

\section{Conclusion and Future Work}
In this project, we try to solve the video generation problem from the video content swap side. 
We propose two methods to train a disentangled frame generator using content and pose information, disentangled encoder-decoder training and CGAN training.
Both auto-ecnoding and CGAN methods can generate video frames with good quality in a small action recognition data set, which support our idea of using disentangled representation for solving video content swap task.
We also found the frame generated using CGAN looks more photo realistic and CGAN appoarch is better at this problem.

Future direction of project is to experiment with the methods on large action data set with various poses, background and different human appearances. 
The large data set will improve the model generalization to the a new pose and content in the test data.
With more complex data set, we would have a more accurate evaluation of the strength and weakness of each method and develop better model for video content swap task.

\bibliography{reference.bib} 

\end{document}